%% file: main.tex
\documentclass{article}

\usepackage{multibib}
\usepackage{microtype}
\usepackage{graphicx}
\usepackage[colorlinks]{hyperref}
\hypersetup{citecolor=blue}

\usepackage{url}
\usepackage{makecell, booktabs}
\usepackage{xcolor}
\usepackage{eqparbox}

\usepackage{microtype}
\usepackage{subcaption}
\usepackage{multirow}
\usepackage{eqparbox}

\usepackage{algorithmic}
\usepackage{nicefrac}
\usepackage{mwe}
\usepackage{icomma}
\usepackage{graphicx}
\usepackage{bm}
\usepackage{amsmath}
\usepackage{amsfonts}
\usepackage[accepted]{icml2020}

\usepackage{silence}
\newsavebox{\tempbox}
\newcolumntype{Y}[1]{
  >{ \begin{lrbox}{\tempbox} }
  c
  <{ \end{lrbox}
     \eqparbox{#1}{\strut\unhcopy\tempbox}
   }
}

\newcolumntype{Q}[1]{>{\begin{lrbox}{\tempbox}}%
    c<{\end{lrbox}\eqparbox{#1}{\unhcopy\tempbox}}}
\newcites{AP}{References}

\icmltitlerunning{Learning What to Defer for Maximum Independent Sets}

\begin{document}

\twocolumn[
\icmltitle{Learning What to Defer for Maximum Independent Sets}

\icmlsetsymbol{equal}{*}

\begin{icmlauthorlist}
\icmlauthor{Sungsoo Ahn}{kaist}
\icmlauthor{Younggyo Seo}{kaist}
\icmlauthor{Jinwoo Shin}{kaist}
\end{icmlauthorlist}

\icmlaffiliation{kaist}{Korea Advanced Institute of Science and Technology (KAIST)}

\icmlcorrespondingauthor{Sungsoo Ahn}{sungsoo.ahn@kaist.ac.kr}

\icmlkeywords{Machine Learning, ICML}

\vskip 0.3in
]

\printAffiliationsAndNotice{}

\begin{abstract}
Designing efficient algorithms for combinatorial optimization appears ubiquitously in various scientific fields. Recently, deep reinforcement learning (DRL) frameworks have gained considerable attention as a new approach: they can automate the design of a solver while relying less on sophisticated domain knowledge of the target problem. However, the existing DRL solvers determine the solution using a number of stages proportional to the number of elements in the solution, which severely limits their applicability to large-scale graphs. In this paper, we seek to resolve this issue by proposing a novel DRL scheme, coined learning what to defer (LwD), where the agent adaptively shrinks or stretch the number of stages by learning to distribute the element-wise decisions of the solution at each stage. We apply the proposed framework to the maximum independent set (MIS) problem, and demonstrate its significant improvement over the current state-of-the-art DRL scheme. We also show that LwD can outperform the conventional MIS solvers on large-scale graphs having millions of vertices, under a limited time budget. 
\end{abstract}

\section{Introduction}\label{sec:intro}
\input{introduction.tex}
\section{Related Works}
\input{related.tex}
\section{Learning What to Defer}
\label{sec:adp}
\input{method.tex}
\section{Experiments}
\input{experiments.tex}

\section{Conclusion}
In this paper, we propose a new deep reinforcement learning scheme for the maximum independent set problem that is scalable to large graphs. Our main contribution is the framework of learning what to defer, which allows the agent to defer the decisions on vertices for efficient expression of complex structures in the solutions. Through extensive experiments, our algorithm shows performance that is both superior to the existing reinforcement learning baseline and competitive with the conventional solvers.

\section*{Acknowledgements}
This work was partly supported by Institute of Information \& communications Technology Planning \& Evaluation (IITP) grant funded by the Korea government (MSIT) (No.~2017-0-01779, A machine learning and statistical inference framework for explainable artificial intelligence). This work was partly supported by Institute of Information \& communications Technology Planning \& Evaluation (IITP) grant funded by the Korea government (MSIT) (No.~2019-0-00075, Artificial Intelligence Graduate School Program (KAIST)). We thank Hyuntak Cha, Hankook Lee, Kimin Lee, Sangwoo Mo, and Jihoon Tak for providing helpful feedbacks and suggestions in preparing the early version of the manuscript.

\bibliography{reference}
\bibliographystyle{icml2020}

\onecolumn
\appendix
\input{appendix.tex}

\newpage
\bibliographyAP{reference}
\bibliographystyleAP{icml2020}

\end{document}

%% file: introduction.tex
Combinatorial optimization is an important mathematical field addressing fundamental questions of computation, where its popular examples include the maximum independent set (MIS, \citealp{miller1960problem}), satisfiability (SAT, \citealp{schaefer1978complexity}) and traveling salesman problem (TSP, \citealp{voigt1831handlungsreisende}). Such problems arise in various applications, e.g., sociology \citep{harary1957procedure}, operations research \citep{feo1994greedy} and bioinformatics \citep{gardiner2000graph}. However, most combinatorial optimization problems are NP-hard and exact solutions are typically intractable to find in practical situations. Over the past decades, researchers have made significant efforts for resolving this issue by designing fast heuristic solvers \citep{knuth1997art, biere2009handbook, mezard2009information} that generate approximate solutions. 

Recently, the remarkable progress in deep learning has stimulated increased interest in learning such heuristics based on deep neural networks (DNNs). Such learning-based approaches are attractive for being able to train a solver on a particular problem while relying less on expert knowledge. As the most straight-forward way, supervised learning schemes train the DNNs to imitate the solutions obtained from existing solvers \citep{vinyals2015pointer, li2018combinatorial, selsam2018learning}. However, the resulting quality and applicability are constrained by those of existing solvers. An ideal direction is to discover new solutions in a fully unsupervised manner, potentially outperforming those based on domain-specific knowledge. 

To this end, recent works \citep{bello2016neural, khalil2017learning, deudon2018learning, kool2018attention} consider using deep reinforcement learning (DRL) based on the Markov decision process (MDP) naturally designed with rewards derived from the optimization objective of the target problem. Then, the corresponding agent can be trained based on existing training schemes of DRL, e.g., Bello et al.~\citep{bello2016neural} trained a TSP solver based on actor-critic framework. Such DRL-based methods are especially attractive since they can even solve unexplored problems where domain knowledge is scarce, and no efficient heuristic is known. 

Unfortunately, the existing DRL-based methods struggle to compete with the existing highly optimized solvers. In particular, the gap becomes significant when the problem requires solutions with higher dimensions. The reasoning is that they mostly emulate greedy algorithms \citep{bello2016neural, khalil2017learning}, i.e., choosing one element of the solution at each stage of MDP. Such a procedure quickly becomes computationally prohibitive for obtaining a large-scale solution. This motivates us to seek for an alternative DRL scheme.

\textbf{Contribution.} 
{
In this paper, we propose a new DRL framework, coined learning what to defer (LwD), for solving large-scale combinatorial optimization problems. Our framework is designed for locally decomposable problems, where the feasibility constraint and the objective can be decomposed by locally connected variables (in a graph). Representative examples of such problems include satisfiability \citep{schaefer1978complexity}, maximum cut (MAXCUT, \citealp{garey1979computers}), and graph coloring \citep{brooks1941colouring}.\footnote{{We note that TSP is not locally decomposable since it has a global constraint of forming a Hamiltonian cycle.}} %Nevertheless, it would be interesting to apply our framework to TSP in the future.}
Among them, we focus on the maximum independent set (MIS) problem; it is a well-studied prototype of NP-hard combinatorial optimization with highly-optimized solvers \citep{andrade2012fast, lamm2017finding, chang2017computing, hespe2019scalable}, where we would like to showcase that our scheme can even compete with existing state-of-the-art solvers. The MIS problem has been used in various applications including classification theory \citep{feo1994greedy}, computer vision \citep{sander2008efficient} and communication \citep{jiang2010distributed}. 
}

%Theoretically, the MIS is impossible to approximate in polynomial time by a constant factor (unless P=NP) \citep{hastad1996clique}, in contrast to (Euclidean or metric) TSP which can be approximated by a factor of $1.5$ \citep{christofides1976worst}. 

%Among them, we focus on the maximum independent set (MIS) problem \citep{miller1960problem} for finding a maximum set of non-adjacent vertices in the graph with the motivation from various applications \citep{feo1994greedy,sander2008efficient,jiang2010distributed} and theoretical hardness \citep{hastad1996clique}. 

%. In particular, LwD is efficient to apply
%for  problems; 
%solving locally decomposable combinatorial problems in large graphs, i.e., we %are concerned with large-scale problems where 
%\footnote{\attention{For completeness, we also provide empirical results on applying LwD to non-MIS problems in Section~\ref{subsec:exp_other}.}}
%Our choice motivates from how the MIS problem has been used in various applications including classification theory \citep{feo1994greedy}, computer vision \citep{sander2008efficient} and communication \citep{jiang2010distributed}. In theory, MIS is impossible to approximate in polynomial time by a constant factor (unless P=NP) \citep{hastad1996clique}, in contrast to (Euclidean or metric) TSP which can be approximated by a factor of $1.5$ \citep{christofides1976worst}.

%The main novelty of LwD is automatically stretching the determination of the solution throughout multiple steps. 
The main novelty of LwD is automatically stretching the determination of the solution throughout multiple steps. In particular, the agent iteratively acts on every undetermined vertex for either (a) determining the membership of the vertex in the solution or (b) deferring the determination to be made in later steps (see Figure~\ref{fig:mdp} for illustration). Inspired by the celebrated survey propagation \citep{braunstein2005survey} for solving the SAT problem, LwD could be interpreted as prioritizing the ``easier'' decisions to be made first, which in turn simplifies the harder ones by eliminating the source of uncertainties. Compared to the greedy strategy \citep{khalil2017learning} which determines the membership of a single vertex at each step, our framework brings significant speedup by learning to make many {element-wise} decisions at once (and deferring the rest).

\input{figure/mdp.tex}

Based on such a speedup, LwD can solve the optimization problem by generating a large number of candidate solutions in a limited time budget, then reporting the best solution among them. For this scenario, it is beneficial for the algorithm to generate diverse candidates. To this end, we additionally give a novel diversification bonus to our agent during training, which explicitly encourages the agent to generate a large variety of solutions. To be specific, we create a ``coupling'' of MDPs to generate two solutions for the given MIS problem and reward the agents for a large deviation between the solutions. The resulting reward efficiently improves the performance at the evaluation.

We empirically validate LwD on various types of graphs including the Erd\"os-R\'enyi \citep{erdHos1960evolution} model, the Barab\'asi-Albert \citep{albert2002statistical} model, the SATLIB \citep{hoos2000satlib} benchmark and real-world graphs. Our algorithm shows consistent superiority over the existing state-of-the-art DRL method \citep{khalil2017learning}. Remarkably, it often outperforms the state-of-the-art MIS solver (KaMIS, \citealp{hespe2019wegotyoucovered}), particularly on large-scale graphs, e.g., in our machine, LwD %running in 276 seconds 
achieves better objectives $3637/276\approx 13$ times faster, compared to KaMIS on the Barab\'asi-Albert graph with two million vertices. Furthermore, we also show that our fully learning-based scheme generalizes well even to graph types unseen during training and works well even for other locally decomposable combinatorial optimization: the maximum weighted independent set problem, the prize collecting maximum independent set problem \citep{hassin2006minimum}, the MAXCUT problem \citep{garey1979computers}, and the maximum-a-posteriori inference problem for the Ising model \citep{onsager1944crystal}. 

%% file: figure/mdp.tex
\begin{figure*}[t]
    \centering
    \includegraphics[width=0.8\textwidth]{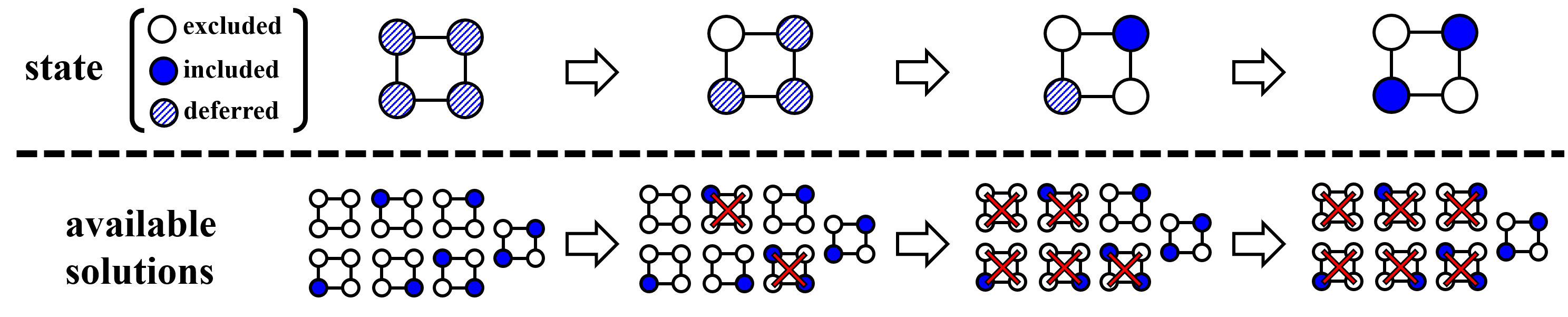}
    \vspace{-0.1in}
    \caption{Illustration of the proposed Markov decision process.}
    \vspace{-0.05in}
    \label{fig:mdp}
\end{figure*}

%% file: related.tex
The maximum independent set (MIS) problem is a prototypical NP-hard task where its optimal solution cannot be approximated by a constant factor in polynomial time unless P = NP \citep{hastad1996clique}.
%\footnote{It is also known to be a $W[1]$-hard problem in terms of fixed-parameter tractability \citep{downey2012parameterized}.} 
Since the problem is NP-hard even to approximate, existing methods \citep{tomita2010simple, san2011exact} for exactly solving the MIS problem suffer from a prohibitive amount of computation in large graphs. To resolve this, researchers have developed a wide range of approximate solvers for the MIS problem (\citealt{andrade2012fast}, \citealt{lamm2017finding}, \citealt{chang2017computing}, \citealt{hespe2019scalable}). Notably, \citet{lamm2017finding} developed a combination of an evolutionary algorithm with graph kernelization techniques for the MIS problem. Later, \citet{chang2017computing} and \citet{hespe2019scalable} further improved the graph kernelization technique by introducing new reduction rules and parallelization based on graph partitioning, respectively. 

In the context of solving combinatorial optimization using neural networks, \citet{hopfield1985neural} first applied the Hopfield-network for solving the traveling salesman problem (TSP). Since then, several works also tried to utilize neural networks in different forms, e.g., see \citet{smith1999neural} for a review of such papers. Such works mostly solve combinatorial optimization through online learning, i.e., training was performed for each problem instance separately. More recently, \citet{vinyals2015pointer} and \citet{bello2016neural} proposed to solve TSP by training attention-based neural networks with offline learning. They showed promising results that stimulated many other works to use neural networks for solving combinatorial problems (\citealt{khalil2017learning}, \citealt{li2018combinatorial}, \citealt{selsam2018learning}, \citealt{deudon2018learning}, \citealt{amizadeh2018learning}, \citealt{kool2018attention}). Importantly, \citet{khalil2017learning} proposed a reinforcement learning framework for solving the minimum vertex cover problem, which is equivalent to solving the MIS problem. They query the agent for each vertex to add as a new member of the vertex cover at each step of the Markov decision process. However, such a greedy procedure hurts the scalability to large-scale graphs, as we mentioned in Section~\ref{sec:intro}. Next, \citet{li2018combinatorial} aim developing a supervised learning framework for solving the MIS problem. At an angle, their framework is similar to ours; they allow stretching the determination of the solution over multiple steps. However, their scheme for stretching the determination is hand-designed and not trainable from data. Furthermore, their scheme requires supervisions, which are (a) highly sensitive to the quality of solvers used for extracting them and (b) often too expensive or almost impossible to obtain. 

%% file: method.tex
In this section, we focus on describing the learning what to defer (LwD) framework for the maximum independent set (MIS) problem. 
Given a graph $\mathcal{G} = (\mathcal{V}, \mathcal{E})$ with vertices $\mathcal{V}$ and edges $\mathcal{E}$, an {\it independent set} is a subset of vertices $\mathcal{I}\subseteq \mathcal{V}$ where no two vertices in the subset are adjacent to each other. A solution to the MIS problem can be represented as a binary vector $\bm{x} = [x_{i} : i \in \mathcal{V}] \in \{0, 1\}^{\mathcal{V}}$ with maximum cardinality $\sum_{i\in\mathcal{V}} x_i$, where each element $x_{i}$ indicates the element-wise decision on the membership of vertex $i$ in the independent set $\mathcal{I}$, i.e., $x_{i} = 1$ if and only if $i \in \mathcal{I}$. Initially, the algorithm has no assumption about its output, i.e., both $x_{i}=0$ and $x_{i}=1$ are possible for all $i\in\mathcal{V}$. At each iteration, the agent acts on each undetermined vertex $i$ by either (a) deciding its membership to be a certain value, i.e., set $x_{i}=0$ or $x_{i}=1$, or (b) deferring the decision to be made later iterations. The agent repeats the action until all the membership of vertices in the independent set is determined.
{One can interpret such a strategy as progressively narrowing down the set of candidate solutions at each iteration (see Figure~\ref{fig:mdp} for illustration). Intuitively, the act of deferring prioritizes to choose the values of the ``easier'' vertices first. After each decision, ``hard'' vertices become easier since decisions on its surrounding easy vertices are better known. } 
%In a sense, this idea resembles how humans guide the behavior of animals, known as \textit{shaping} \citep{peterson2004day}, where the easier concepts are fixed first for the animals to affect their decisions on harder concept. We additionally provide illustration of the algorithm in Appendix~\ref{sec:app_graph}.

We note that LwD is applicable to combinatorial optimization problems other than the MIS problem, e.g., by considering the decision of non-binary vector $\bm{x}$. Especially, the LwD framework is attractive to apply for locally decomposable combinatorial problems, i.e., problems where the feasibility constraints are decomposable by locally connected variables. Popular problems such as satisfiability \citep{schaefer1978complexity}, maximum cut \citep{garey1979computers}, and graph coloring \citep{brooks1941colouring} also fall into this category. For these problems, making multiple ``local'' decisions on variables at once likely does not violate the feasibility, and this is what LwD needs for its efficient implementation. 

%\attention{
%The rest of this section is organized as follows. In Section \ref{subsec:mdp}, we describe the deferred Markov decision process (MDP) used for our algorithm. Next, in Section \ref{subsec:diversification}, we propose an additional diversification bonus reward that could be used in conjunction with the proposed MDP to further boost the performance of our algorithm. We then provide description of the neural architecture and the proximal policy optimization method used for our algorithm in Section \ref{subsec:ppo}. 
%}

\input{figure/hamming.tex}

\subsection{Deferred Markov Decision Process}
\label{subsec:mdp}
We formulate the proposed algorithm as a pair of a MDP and an agent, i.e., a policy. At a high level, the MDP initializes its states on the given graph and generates a solution at termination. We further decompose the MIS objective into a summation of rewards distributed over the MDP.

%We train the agent to maximize the MIS objective, formulated as a cumulative sum of rewards over the MDP.

{\bf State.} 
Each state of the MDP is represented as a \textit{vertex-state} vector $\bm{s} = [s_{i}: i \in \mathcal{V}] \in \{0, 1, \ast\}^{\mathcal{V}}$, where the vertex $i\in \mathcal{V}$ is determined to be excluded or included in the independent set whenever $s_{i}=0$ or $s_{i}=1$, respectively. Otherwise, $s_{i}=\ast$ indicates that the determination has been deferred and expected to be made in later iterations. The MDP is initialized with the deferred vertex-states, i.e., $s_{i}=\ast$ for all $i \in \mathcal{V}$, and terminated when (a) there is no deferred vertex-state left or (b) time limit is reached.

{\bf Action.} 
Actions correspond to new assignments for the next state of vertices. Since vertex-states of included and excluded vertices are immutable, the assignments are defined only on the deferred vertices. It is represented as a vector $\bm{a}_{\ast} = [a_{i}: i \in \mathcal{V}_{\ast}] \in \{0, 1, \ast\}^{\mathcal{V}_{\ast}}$ where $\mathcal{V}_{\ast}$ denotes a set of current deferred vertices, i.e., $\mathcal{V}_{\ast} = \{i: i \in \mathcal{V}, x_{i} = \ast\}$.

{\bf Transition.}
Given two consecutive states $\bm{s}, \bm{s}^{\prime}$ and the corresponding assignment $\bm{a}_{\ast}$, the transition $P_{\bm{a}_{\ast}}(\bm{s}, \bm{s}^{\prime})$ consists of two deterministic phases: the \textit{update phase} and the \textit{clean-up phase}.  The update phase takes account of the assignment $\bm{a}_{\ast}$ generated by the policy for the corresponding vertices $\mathcal{V}_{\ast}$ to result in an intermediate vertex-state $\widehat{\bm{s}}$, i.e., $\widehat{s}_{i} = a_{i}$ if $i \in \mathcal{V}_{\ast}$ and $\widehat{s}_{i} = s_{i}$ otherwise. The clean-up phase modifies the intermediate vertex-state vector $\widehat{\bm{s}}$ to yield a valid vertex-state vector $\bm{s}^{\prime}$, where the included vertices are only adjacent to the excluded vertices. To this end, whenever there exists a pair of included vertices adjacent to each other, they are both mapped back to the deferred vertex-state. Next, the MDP excludes any deferred vertex neighboring with an included vertex.\footnote{We note that such a clean-up phase can be replaced by training the agent with a soft penalty for solutions corresponding to an invalid independent set. In our experiments, such an algorithm also performs well with only marginal degradation in its performance.} See Figure~\ref{subfig:transition} for a more detailed illustration of the transition between two states.

When the MDP makes all the determination, i.e., at termination, one can (optionally) improve the determined solution by applying the $2$-improvement local search algorithm \citep{feo1994greedy, andrade2012fast}; it increases the size of the independent set greedily by removing one vertex and adding two vertices until no modification is possible.

{\bf Reward.}
Finally, we define the \textit{cardinality reward} $R(\bm{s}, \bm{s}^{\prime})$ as the increase in cardinality of included vertices. To be specific, we define it as $R(\bm{s}, \bm{s}^{\prime}) = \sum_{i\in\mathcal{V}_{\ast} \backslash \mathcal{V}_{\ast}^{\prime}} s^{\prime}_{i}$, where $\mathcal{V}_{\ast}$ and $\mathcal{V}_{\ast}^{\prime}$ are the set of vertices with deferred vertex-state with respect to $\bm{s}$ and $\bm{s}^{\prime}$, respectively. By doing so, the overall reward of the MDP corresponds to the cardinality of the independent set returned by our algorithm. 

\subsection{Diversification Reward}
\label{subsec:diversification}
Next, we introduce an additional {\it diversification reward} for encouraging diversification of solutions generated by the agent. Such a regularization is motivated by our evaluation method, which samples multiple candidate solutions to report the best one as the final output. For such scenarios, it would be beneficial to generate diverse solutions of a high maximum score, rather than ones of similar scores. {
To this end, we ``couple'' two copies of MDPs defined on an identical graph $\mathcal{G}$ (as in Section~\ref{subsec:mdp}) into a new MDP. Then the new MDP is associated with a pair of distinct vertex-state vectors $(\bm{s}, \bar{\bm{s}})$ from the coupled MDPs. Furthermore, the corresponding agents work independently to result in a pair of solutions $(\bm{x}, \bar{\bm{x}})$.
} Then, we directly reward the deviation between the coupled solutions in terms of $\ell_{1}$-norm, i.e., $\lVert \bm{x} - \bar{\bm{x}} \rVert_{1}$. To be specific, the deviation is decomposed into rewards in each iteration of the MDP defined by %as follows: 
\begin{equation*}
R_{\text{div}}(\bm{s}, \bm{s}^{\prime}, \bar{\bm{s}}, \bar{\bm{s}}^{\prime})=\sum_{i \in \widehat{\mathcal{V}}} | s_{i}^{\prime} - \bar{s}_{i}^{\prime} |, 
\end{equation*} where $\widehat{\mathcal{V}} = (\mathcal{V}_{*} \setminus \mathcal{V}^{\prime}_{*}) \cup (\bar{\mathcal{V}}_{*} \setminus \bar{\mathcal{V}}^{\prime}_{*})$ and $(\bm{s}^{\prime}, \bar{\bm{s}}^{\prime})$ denotes the next pair of vertex-states in the coupled MDP. Note that $\widehat{\mathcal{V}}$ indicates the most recently updated vertices in each MDP. In practice, such a reward $R_{\text{div}}$ can be used along with the maximum entropy regularization to achieve the best performance. See Figure~\ref{subfig:hamming} for an example of coupled MDP with the proposed reward. 
\input{table/small_scale.tex}

{We remark that the entropy regularization \citep{williams1991function} of the policy plays a similar role (of encouraging exploration) to our diversification reward. However, the entropy regularization attempts to generate diverse trajectories of the same MDP, which does not necessarily lead to diverse solutions at last, since there exist many trajectories resulting in the same solution (see Section~\ref{subsec:mdp}). We instead directly maximize the diversity among solutions by a new reward term.
}

\subsection{Training with Proximal Policy Optimization}
\label{subsec:ppo}
Our algorithm is based on actor-critic training with policy network $\pi(\bm{a} | \bm{s})$ and value network $V(\bm{s})$ following the GraphSAGE architecture \citep{hamilton2017inductive}. Note that we only consider the subgraph that is induced on the deferred vertices $\mathcal{V}_{\ast}$ as the input of the networks since the determined part of the graph no longer affects the future rewards of the MDP. For a detailed explanation of the architectures for the policy and the value networks, see Appendix~\ref{sec:app_detail}. We utilize vertex degrees and the current iteration-index of the MDP as the input features of the neural network. To train the agent, we use the proximal policy optimization \citep{schulman2017proximal}. To be specific, we train the networks to maximize the following objective:
\begin{align*}
\mathcal{L} :&= \mathbb{E}_{t}\bigg[\min\bigg(A^{(t)}\prod_{i \in \mathcal{V}} r_{i}^{(t)},A^{(t)}\prod_{i \in \mathcal{V}} \widetilde{r}_{i}^{(t)}\bigg)\bigg], \\
A^{(t)} &= \sum_{\ell=0}^{T-t}{\big(R^{(t+\ell)} + \alpha R^{(t+\ell)}_{\text{div}}\big)} - V(\bm{s}^{(t)}), \\
r_{i}^{(t)} &= \frac{\pi(a^{(t)}_{i} | \bm{s}^{(t)})}{\pi_{\text{old}}(a^{(t)}_{i} | \bm{s}^{(t)})}, \quad 
\widetilde{r}_{i}^{(t)} = \mathtt{clip}(r_{i}^{(t)}, 1-\varepsilon, 1+\varepsilon),
\end{align*}
where $\bm{s}^{(t)}$, $\bm{a}^{(t)}$, $R^{(t)}$ and $R^{(t)}_{\text{div}}$ denotes the $t$-th vertex-state vector, action vector, cardinality reward, and diversification reward, respectively. In addition,  ${\pi_{\text{old}}}$ is the policy network with parameters from the previous iteration of updates, $T$ is the maximum number of steps for the MDP, and $\alpha \geq 0$ is the hyperparameter to be tuned. The clipping function $\mathtt{clip}(r, r_{\mathtt{min}}, r_{\mathtt{max}})$ projects the ratio of action probabilities $r$ into an interval $[r_{\mathtt{min}}, r_{\mathtt{max}}]$ for conservatively updating the agent. Note that the clipping is applied for each vertex, unlike the original framework where clipping is applied once \citep{schulman2017proximal}.

%% file: figure/hamming.tex
\begin{figure*}[t]
    \begin{minipage}{0.50\textwidth}
    \centering
    \includegraphics[width=0.8\textwidth]{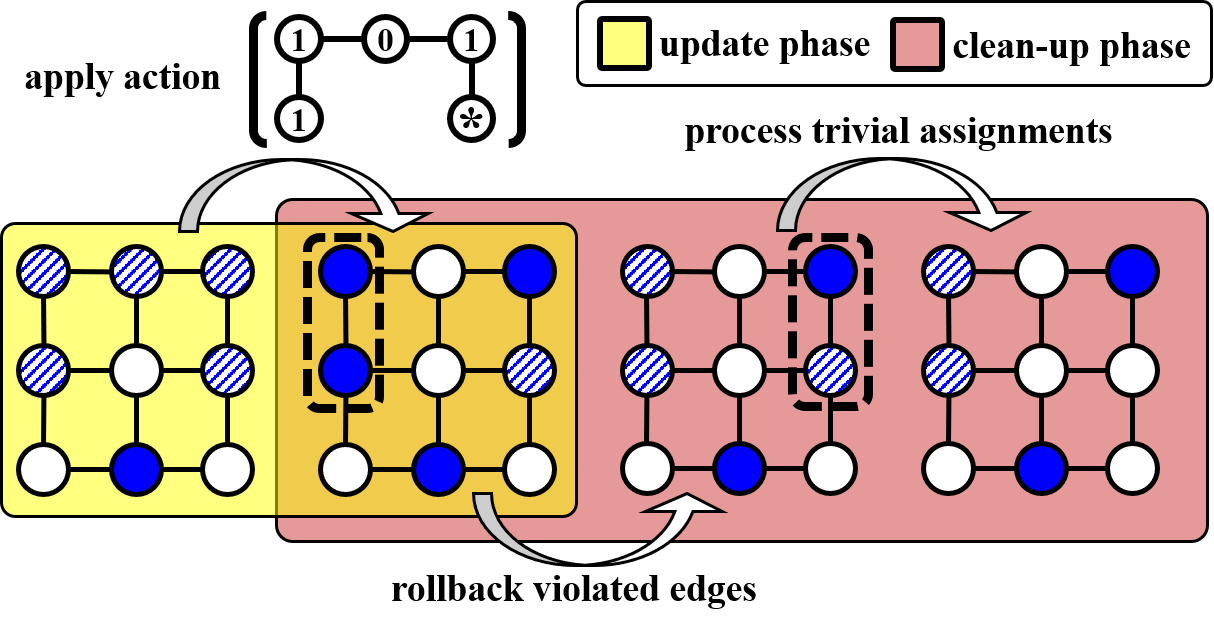}
    \caption{Illustration of the transition function.}
    \label{subfig:transition}
    \end{minipage}
    \hfill
    \begin{minipage}{0.44\textwidth}
    \centering
    \includegraphics[width=0.8\textwidth]{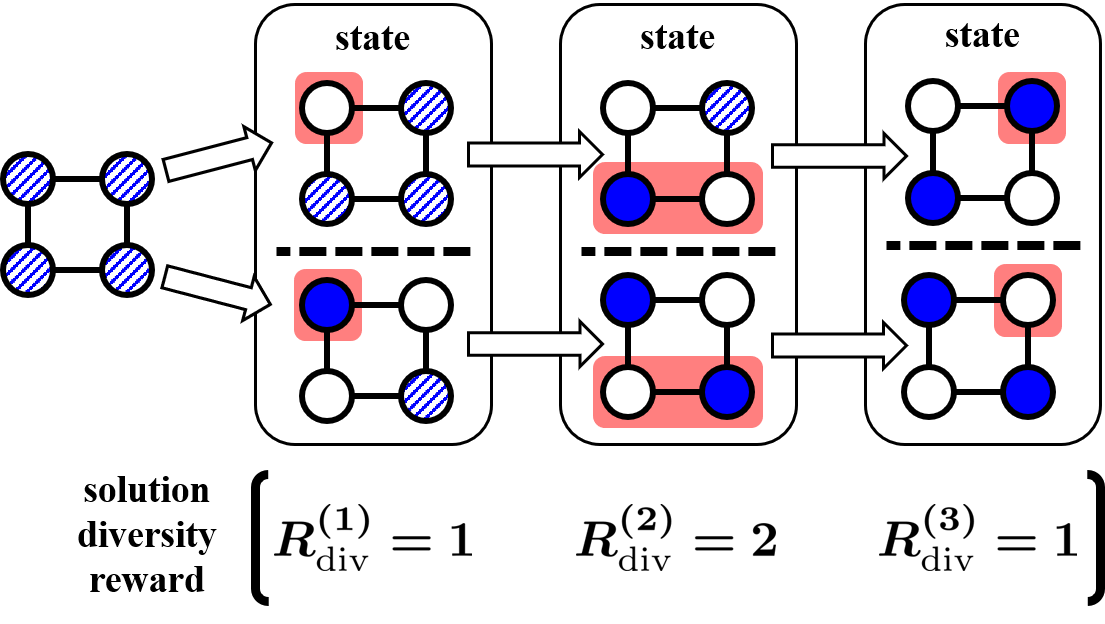}
    \caption{Illustration of the solution diversity reward.}
    \label{subfig:hamming}
    \end{minipage}
    \hfill
\end{figure*}

%% file: table/small_scale.tex
\begin{table*}[t!]
\centering
\caption{{
Objectives achieved from the MIS solvers on moderately size dataset, 
where the best objectives are marked in bold. 
Running times (in seconds) of the deep-learning based algorithms are 
provided in brackets.
When running CPLEX and KaMIS, time limits were set by $5$ and $30$ seconds for 
$\{$ER, BA, HK, WS$\}$ and $\{$SATLIB, PPI, REDDIT, as-Caida$\}$ datasets, 
respectively, so that they spend more (or comparable) time than LwD and LwD$^{\dagger}$ to find their solutions.
The minimum and the maximum number of vertices in each dataset are denoted by 
$N_{\mathtt{min}}$ and $N_{\mathtt{max}}$, respectively. 
}
}
\vspace{0.1in}
\label{tab:perf_synthetic}
\scalebox{0.7}{
\begin{tabular}{
c
c
c
c
@{\hspace{0.0cm}}
c
c
c@{\hspace{0.1cm}}
c
c
c@{\hspace{0.1cm}}
c
c
c
} 
\toprule[1.2pt] 
&
&
&
& \multicolumn{2}{c}{Classic}
&
& \multicolumn{2}{c}{SL-based}
&
& \multicolumn{3}{c}{RL-based}
\\
\cmidrule[1.2pt]{5-6}
\cmidrule[1.2pt]{8-9} 
\cmidrule[1.2pt]{11-13}
\multicolumn{1}{c}{Type}
& \multicolumn{1}{c}{$N_{\mathtt{min}}$}
& \multicolumn{1}{c}{$N_{\mathtt{max}}$}
&
& CPLEX
& KaMIS
&
& TGS
& TGS$^\dagger$
& 
& S2V-DQN
& LwD
& LwD$^\dagger$
\\
\midrule[1.2pt]
\multirow{3.5}{*}{ER}
&\phantom{00}50
&\phantom{00~}100
&
&\phantom{0~0}\textbf{21.11}
&\phantom{0~0}\textbf{21.11}
&
&\phantom{0~0}19.90 \small{(0.32)\phantom{0}}
&\phantom{0~0}\textbf{21.11} \small{(0.65)\phantom{0}}
&
&\phantom{0~0}20.61 \small{(0.03)\phantom{0}}
&\phantom{0~0}21.04 \small{(0.01)\phantom{0}}
&\phantom{0~0}\textbf{21.11} \small{(0.15)\phantom{0}}
\\
\addlinespace
&\phantom{0}100
&\phantom{00~}200
&
&\phantom{0~0}27.87
&\phantom{0~0}\textbf{27.95}
&
&\phantom{0~0}24.94 \small{(1.46)\phantom{0}}
&\phantom{0~0}\textbf{27.95} \small{(1.54)\phantom{0}}
&
&\phantom{0~0}26.27 \small{(0.08)\phantom{0}}
&\phantom{0~0}27.67 \small{(0.03)\phantom{0}}
&\phantom{0~0}\textbf{27.95} \small{(0.61)\phantom{0}}
\\
\addlinespace
&\phantom{0}400
&\phantom{00~}500
&
&\phantom{0~0}31.73
&\phantom{0~0}39.61
&
&\phantom{0~0}33.46 \small{(0.93)\phantom{}}
&\phantom{0~0}39.43 \small{(12.37)\phantom{}}
&
&\phantom{0~0}35.05 \small{(0.63)\phantom{0}}
&\phantom{0~0}38.29 \small{(0.16)\phantom{0}}
&\phantom{0~0}\textbf{39.81} \small{(5.91)\phantom{0}}
\\
\midrule[1.2pt]
\multirow{3.5}{*}{BA} 
&\phantom{00}50
&\phantom{00~}100
&
&\phantom{0~0}\textbf{32.07}
&\phantom{0~0}\textbf{32.07}
&
&\phantom{0~0}31.77 \small{(0.24)\phantom{0}}
&\phantom{0~0}\textbf{32.07} \small{(0.25)\phantom{0}}
&
&\phantom{0~0}31.96 \small{(0.02)\phantom{0}}
&\phantom{0~0}\textbf{32.07} \small{(0.01)\phantom{0}}
&\phantom{0~0}\textbf{32.07} \small{(0.11)\phantom{0}}
\\
\addlinespace
&\phantom{0}100
&{\phantom{00~}200}
&
&\phantom{0~0}\textbf{66.07}
&\phantom{0~0}\textbf{66.07}
&
&\phantom{0~0}65.25 \small{(0.33)}\phantom{0}
&\phantom{0~0}\textbf{66.07} \small{(0.52)}\phantom{0}
&
&\phantom{0~0}65.52 \small{(0.05)}\phantom{0}
&\phantom{0~0}66.05 \small{(0.01)}\phantom{0}
&\phantom{0~0}\textbf{66.07} \small{(0.22)}\phantom{0}
\\
\addlinespace
&\phantom{0}400
&\phantom{00~}500
&
&\phantom{00~}\textbf{204.1} 
&\phantom{00~}\textbf{204.1}
&
&\phantom{00~}201.2 \small{(0.72)}\phantom{0}
&\phantom{00~}\textbf{204.1} \small{(7.86)}\phantom{0}
&
&\phantom{00~}202.9 \small{(0.18)}\phantom{0}
&\phantom{00~}204.0 \small{(0.02)}\phantom{0}
&\phantom{00~}\textbf{204.1} \small{(0.87)}\phantom{0}
\\
\midrule[1.2pt]
\multirow{3.5}{*}{HK} 
&\phantom{00}50
&\phantom{00~}100
&
&\phantom{0~0}\textbf{23.95}
&\phantom{0~0}\textbf{23.95}
&
&\phantom{0~0}23.39 \small{(0.29)}\phantom{0}
&\phantom{0~0}\textbf{23.95} \small{(0.60)}\phantom{0}
&
&\phantom{0~0}23.77 \small{(0.03)}\phantom{0}
&\phantom{0~0}\textbf{23.95} \small{(0.02)}\phantom{0}
&\phantom{0~0}\textbf{23.95} \small{(0.15)}\phantom{0}
\\
\addlinespace
&\phantom{0}100
&{\phantom{00~}200}
&
&\phantom{0~0}\textbf{50.15}
&\phantom{0~0}\textbf{50.15}
&
&\phantom{0~0}48.74 \small{(0.43)}\phantom{0}
&\phantom{0~0}\textbf{50.15} \small{(2.00)}\phantom{0}
&
&\phantom{0~0}49.64 \small{(0.05)}\phantom{0}
&\phantom{0~0}50.12 \small{(0.04)}\phantom{0}
&\phantom{0~0}\textbf{50.15} \small{(0.34)}\phantom{0}
\\
\addlinespace
&\phantom{0}400
&\phantom{00~}500
&
&\phantom{00~}156.8
&\phantom{00~}\textbf{157.0}
&
&\phantom{00~}152.1 \small{(0.92)}\phantom{0}
&\phantom{00~}\textbf{157.0} \small{(7.63)}\phantom{0}
&
&\phantom{00~}152.8 \small{(0.22)}\phantom{0}
&\phantom{00~}156.8 \small{(0.14)}\phantom{0}
&\phantom{00~}\textbf{157.0} \small{(1.63)}\phantom{0}
\\
\midrule[1.2pt]
\multirow{3.5}{*}{WS} 
&\phantom{00}50
&\phantom{00~}100
&
&\phantom{0~0}\textbf{23.08}
&\phantom{0~0}\textbf{23.08}
&
&\phantom{0~0}21.90 \small{(0.34)}\phantom{0}
&\phantom{0~0}\textbf{23.08} \small{(0.71)}\phantom{0}
&
&\phantom{0~0}22.64 \small{(0.03)}\phantom{0}
&\phantom{0~0}23.07 \small{(0.03)}\phantom{0}
&\phantom{0~0}\textbf{23.08} \small{(0.11)}\phantom{0}
\\
\addlinespace
&\phantom{0}100
&{\phantom{00~}200}
&
&\phantom{0~0}47.17
&\phantom{0~0}\textbf{47.18}
&
&\phantom{0~0}44.55 \small{(0.49)}\phantom{0}
&\phantom{0~0}\textbf{47.18} \small{(1.89)}\phantom{0}
&
&\phantom{0~0}45.39 \small{(0.06)}\phantom{0}
&\phantom{0~0}47.11 \small{(0.06)}\phantom{0}
&\phantom{0~0}\textbf{47.18} \small{(0.23)}\phantom{0}
\\
\addlinespace
&\phantom{0}400
&\phantom{00~}500
&
&\phantom{00~}138.3
&\phantom{00~}\textbf{143.3}
&
&\phantom{00~}134.8 \small{(1.15)}\phantom{0}
&\phantom{00~}143.2 \small{(6.08)}\phantom{0}
&
&\phantom{00~}132.2 \small{(0.23)}\phantom{0}
&\phantom{00~}142.1 \small{(0.17)}\phantom{0}
&\phantom{00~}\textbf{143.3} \small{(0.90)}\phantom{0}
\\
\midrule[1.2pt]
{SATLIB}
&{1209}
&{\phantom{0~}1347}
&
&\phantom{00~}426.8
&\phantom{00~}\textbf{426.9}
&
&\phantom{00~}418.1 \small{(19.6)}\phantom{0}
&\phantom{00~}426.7 \small{(63.0)}\phantom{0}
&
&\phantom{00~}413.8 \small{(2.3)}\phantom{00}
&\phantom{00~}424.8 \small{(1.8)}\phantom{00}
&\phantom{00~}426.7 \small{(7.1)}\phantom{00}
\\
\addlinespace
{PPI}
&{\phantom{0}591}
&{\phantom{0~}3480}
&
&\phantom{00~.}\textbf{1148}
&\phantom{00~.}\textbf{1148}
&
&\phantom{00~.}1128 \small{(20.9)}\phantom{0}
&\phantom{00~.}\textbf{1148} \small{(568.9)}\phantom{}
&
&\phantom{000~.}893 \small{(6.3)}\phantom{00}
&\phantom{00~.}1147 \small{(1.8)}\phantom{00}
&\phantom{00~.}\textbf{1148} \small{(30.8)}\phantom{0}
\\
\addlinespace
REDDIT-M-5K
&\phantom{00}22
&\phantom{0~}3648
&
&\phantom{00~}\textbf{370.6}
&\phantom{00~}\textbf{370.6}
&
&\phantom{00~}367.1 \small{(0.88)}\phantom{0}
&\phantom{00~}\textbf{370.6} \small{(1.7)}\phantom{00}
&
&\phantom{00~}370.1 \small{(0.1)}\phantom{00}
&\phantom{00~}\textbf{370.6} \small{(0.6)}\phantom{00}
&\phantom{00~}\textbf{370.6} \small{(2.0)}\phantom{00}
\\
\addlinespace
REDDIT-M-12K
&{\phantom{000}2}
&{\phantom{0~}3782}
&
&\phantom{00~}\textbf{303.5}
&\phantom{00~}\textbf{303.5}
&
&\phantom{00~}300.5 \small{(0.75)}\phantom{0}
&\phantom{00~}\textbf{303.5} \small{(22.1)}\phantom{0}
&
&\phantom{00~}302.8 \small{(1.9)}\phantom{00}
&\phantom{00~}292.6 \small{(0.1)}\phantom{00}
&\phantom{00~}\textbf{303.5} \small{(2.0)}\phantom{00}
\\
\addlinespace
REDDIT-B
&{\phantom{000}6}
&{\phantom{0~}3782}
&
&\phantom{00~}\textbf{329.3}
&\phantom{00~}\textbf{329.3}
&
&\phantom{00~}327.6 \small{(0.7)}\phantom{00}
&\phantom{00~}\textbf{329.3} \small{(2.5)}\phantom{00}
&
&\phantom{00~}328.6 \small{(0.1)}\phantom{00}
&\phantom{00~}\textbf{329.3} \small{(0.2)}\phantom{00}
&\phantom{00~}\textbf{329.3} \small{(3.0)}\phantom{00}
\\
\addlinespace
{as-Caida}
&{8020}
&{26 475}
&
&\phantom{0.}\textbf{20 049}
&\phantom{0.}\textbf{20 049}
&
&\phantom{0.}19 921 \small{(65.85)}\phantom{}
&\phantom{0.}\textbf{20 049} \small{(601.4)}\phantom{}
&
&\phantom{000~.}324 \small{(34.8)}\phantom{0}
&\phantom{0.}\textbf{20 049} \small{(6.1)}\phantom{00}
&\phantom{0.}\textbf{20 049} \small{(34.6)}\phantom{0}
\\
\bottomrule[1.2pt]
\end{tabular}
}
\end{table*}

%% file: experiments.tex
\label{sec:exp}

\input{table/large_scale.tex}
\input{figure/tradeoff.tex}

In this section, we report experimental results on the proposed learning what to defer (LwD) framework described in Section~\ref{sec:adp} for solving the maximum independent set (MIS) problem. {To this end, we evaluate our framework with and without the local search element described in Section~\ref{subsec:mdp}; we coin the algorithms by LwD$^{\dagger}$ and LwD, respectively.} Note that we also include the evaluations of our framework on other locally decomposable combinatorial problems to demonstrate its potential for being applied to a broader domain (See Section~\ref{subsec:exp_other}). We perform every experiment using a single GPU (NVIDIA RTX 2080Ti) and a single CPU (Intel Xeon E5-2630 v4). For a more detailed description for the experiments, see Appendix~\ref{sec:app_detail}.

\textbf{Baselines.} 
For comparison with the deep learning-based methods, we consider the deep reinforcement learning (DRL) framework by \citet{khalil2017learning}, coined S2V-DQN, and supervised learning (SL) framework by \citet{li2018combinatorial}, coined TGS. We also consider its variant, coined TGS$^{\dagger}$, equipped with additional graph reduction and local search algorithms. We remark that other existing deep learning-based schemes for solving combinatorial optimization, e.g., works done by \citet{bello2016neural} and \citet{kool2018attention}, are not comparable since they propose a neural architecture specialized to TSP-like problems.

We additionally consider two conventional MIS solvers as competitors. First, we consider the integer programming solver IBM ILOG CPLEX Optimization Studio V12.9.0 \citep{ilog2014cplex}, coined CPLEX. We also consider the MIS solver based on the recently developed techniques \citep{lamm2015graph, lamm2017finding, hespe2019scalable}, coined KaMIS, which won the PACE 2019 challenge at the vertex cover track \citep{hespe2019wegotyoucovered}. 

We remark that comparisons among the algorithms should be made carefully by accounting for the scope of applications concerning each algorithm. In particular, KaMIS, TGS$^{\dagger}$ and LwD$^{\dagger}$, rely on heuristics specialized for the MIS problem, e.g., the local search algorithm, while CPLEX, S2V-DQN, and LwD can be applied even to non-MIS problems. The integer programming solver CPLEX can provide the proof of optimality in addition to its solution. Furthermore, TGS and TGS$^{\dagger}$ are only applicable to problems where solvers are available for obtaining supervised solutions.

\textbf{Datasets.} 
Experiments were conducted on a broad range of graphs to include both real-world and large-scale graphs. First, we consider random graphs generated from models designed to imitate the characteristics of real-world graphs. Specifically, we consider the models proposed by Erd\"os-R\'enyi (ER, \citealp{erdHos1960evolution}), Barab\'asi-Albert (BA, \citealp{albert2002statistical}), Holme and Kim (HK, \citealp{holme2002growing}), and Watts-Strogatz (WS, \citealp{watts1998collective}). For convenience of notation, the synthetic datasets are specified by their type of generative model and size, e.g., ER-$[N_{\mathtt{min}}, N_{\mathtt{max}}]$ denotes the set of ER graphs generated with the number of vertices uniformly sampled from the interval $[N_{\mathtt{min}}, N_{\mathtt{max}}]$.
Next, we consider real-world graph datasets, namely the SATLIB, PPI, REDDIT, as-Caida, Citation, Amazon, and Coauthor datasets constructed from the SATLIB benchmark \citep{hoos2000satlib}, protein-protein interactions \citep{hamilton2017inductive}, social networks \citep{yanardag2015deep}, road networks \citep{leskovec2016snap}, co-citation network \citep{yang2016revisiting}, co-purchasing network \citep{mcauley2015image}, and academic relationship network \citep{sen2008collective}, respectively. 

\subsection{Performance Evaluation}
\label{sec:perf_eval}
We first show the performance of our algorithm along with other baselines for the MIS problem. 

\input{table/generalization.tex}
\input{figure/ablation.tex}

\textbf{Moderately sized graphs.}
First, we provide the experimental results for the datasets with the number of vertices up to $30~000$.%\footnote{Khalil et al.~\citep{khalil2017learning} only reported results from training on graphs with the number of vertices up to $500$ vertices.} 
In Table~\ref{tab:perf_synthetic}, we observe that LwD$^{\dagger}$ consistently achieves the best objective except for the SATLIB dataset. Surprisingly, it even outperforms CPLEX and KaMIS on some datasets, e.g., LwD$^{\dagger}$ achieves a strictly better objective than every other baseline on the ER-$[400, 500]$ dataset. Next, LwD significantly outperforms its DL competitors, i.e., S2V-DQN and TGS, for datasets except for the REDDIT-M-12K dataset. The gap grows for the large-scale dataset, e.g., S2V-DQN underperforms significantly for the as-Caida dataset.

\textbf{Million-scale graphs.}
Next, we highlight the scalability of our framework by evaluating the algorithms under the million-scale graphs. To this end, we generate synthetic graphs from the BA, SW, and WS models with half, one, and two million vertices.\footnote{The ER model requires a computationally prohibitive amount of memory for the large-scale graph generation.} We use the networks trained on the $\{$BA, HK, WS$\}$-$[400, 500]$ datasets for evaluation on graphs from the same generative model. We run CPLEX and KaMIS with a time limit of $1000$ seconds so that they spend more time than our methods to find the solutions.\footnote{However, the solvers consistently violate the time limit due to their expensive pre-solving process.} Furthermore, we exclude S2V-DQN from comparison for being computationally infeasible to be evaluated on such large-scale graphs. 

In Table~\ref{tab:large_synth}, we observe that LwD and LwD$^\dagger$ mostly outperform other algorithms, i.e., they achieve a better objective in a shorter amount of time. In particular, LwD achieves a higher objective than KaMIS in BA graph with two million vertices along with approximately $\times 13$ speedup. The only exception is the WS graph with five hundred thousand vertices, where LwD and LwD$^{\dagger}$ achieve smaller objective than KaMIS, while still being much faster. Such a result highlights the scalability of our algorithm. It is somewhat surprising to observe that LwD achieves objectives similar to LwD$^{\dagger}$, while TGS underperforms significantly compared to TGS$^{\dagger}$. This validates that our method is not sensitive to using the existing heuristics for achieving high performance, unlike the previous method.

\textbf{Trade-offs between objective and time.}
We further investigate the trade-offs between objective and time for the considered algorithms. To this end, we evaluate algorithms on ER-$[400, 500]$ and SATLIB datasets with varying numbers of samples or time limits. In Figure~\ref{fig:tradeoff}, it is remarkable to observe that both LwD and LwD$^\dagger$ achieves a better objective than the CPLEX solver on both datasets under reasonably limited time. Furthermore, LwD and LwD$^\dagger$ outperforms KaMIS for running time smaller than $10$ and $20$ seconds, respectively.

\textbf{Generalization capability.}
Finally, we examine the potential of the deep learning-based solvers as generic solvers, i.e., whether the solvers generalize well to graph types unseen during training. To this end, we evaluate LwD, LwD$^{\dagger}$, TGS, TGS$^{\dagger}$, and S2V-DQN on Citation, Amazon and Coauthor graph datasets. In Table~\ref{tab:gen_real}, we observe that LwD$^\dagger$ achieves near-optimal objective for all of the considered graphs, despite being trained on graphs with different type and a smaller number of vertices. On the other side, S2V-DQN underperforms significantly compared to LwD, i.e., it achieves worse approximation ratios while being slower. We also observe that TGS achieves similar approximation ratios compared to LwD$^\dagger$ but takes a longer time in our experimental setups.

\input{table/other_comb.tex}

\subsection{Ablation Study} 
We now ablate each component of our algorithm to validate its effectiveness. First, we confirm that stretching the determination process indeed improves the performance of LwD. Then we show that the effectiveness of solution diversification reward.

\textbf{Deferring the decision.} 
We first show that ``deferring'' the decision for the MIS problem indeed helps to solve the MIS problem. Specifically, we experiment with varying the maximum number of iterations $T$ in MDP by $T \in \{2, 4, 8, 16, 32\}$ on ER-$(50, 100)$ dataset. Figure~\ref{fig:ablation_time} reports the corresponding training curves. We observe that the performance of LwD improves whenever the agent receives more time to generate the final solution, which verifies that the deferring the decisions plays a crucial role in solving the MIS problem.

\textbf{Solution diversification reward.} 
Next, we inspect the contribution of the solution diversity reward used in our algorithm. To this end, we trained agents with four options: (a) without any exploration bonus, coined Base, (b) with the conventional entropy bonus \citep{williams1991function}, coined Entropy, (c) with the proposed diversification bonus, coined Diverse, and (d) with both of the bonuses, coined Entropy$+$Diverse. Figure~\ref{fig:ablation_hamming} demonstrates the corresponding training curves for validation scores. We observe that the agent trained with the proposed diversification bonus outperforms other agents in terms of validation score, confirming the effectiveness of our proposed reward. In addition, combining both methods, i.e., Entropy$+$Diverse, achieves the best performance.

Finally, we verify our claim that the maximum entropy regularization fails to capture the diversity of solutions effectively, while the proposed solution diversity reward term does. To this end, we compare the fore-mentioned agents with respect to the $\ell_{1}$-deviations between the coupled intermediate vertex-states $\bm{s}$ and $\bar{\bm{s}}$, defined as $|\{i: i \in \mathcal{V}, s_{i} \neq \bar{s}_{i}\}|/|\mathcal{V}|$. We show the corresponding results in Figure~\ref{fig:ablation_dev}. We observe that the entropy regularization promotes large deviation during the intermediate stages, but converges to solutions with smaller deviation. On the contrary, agents trained on diversification rewards succeed in enlarging the deviation between the final solutions. 

\subsection{Other Combinatorial Optimization{s}}
\label{subsec:exp_other}
{Now we evaluate our framework on other locally decomposable combinatorial optimization problems, the maximum weighted independent set problem (MWIS), the prize collecting maximum independent set problem (PCMIS,  \citealp{hassin2006minimum}), the maximum cut problem (MAXCUT, \citealp{garey1979computers}) and the maximum-a-posteriori inference problem for the Ising models \citep{onsager1944crystal}. We compare our algorithm to CPLEX, which is also capable of solving the considered problems with its integer programming framework. }
%\attention{Here, our goal is to demonstrate the generalizability of our LwD framework to various problems without relying on domain-secific knowledge, instead of showing competitive performance compared to the existing state-of-the-art solvers (as we did for the MIS problem).} 
See Appendix~\ref{sec:app_variant} for detailed descriptions of the problems. 

We report the results of the experiments conducted on the ER datasets in Table~\ref{tab:mwis}. Surprisingly, we observe that LwD outperforms CPLEX for most problems and graphs under the time limit of $5$ seconds. In particular, CPLEX fails to produce reasonable solutions for the PCMIS and the Ising problems for ER-$[100, 200]$ and ER-$[400, 500]$ graphs on the PCMIS, MAXCUT and the Ising problems. This is because the CPLEX solves the PCMIS, MAXCUT and the Ising problems by hard integer quadratic programming (IQP). Such an issue does not exist in the MIS and MWIS problems as CPLEX solves them by integer linear programming (ILP), which is easier to solve. The proposed LwD framework does not rely on such domain-specific knowledge of IQP vs.\ ILP, and is more robust under various problems in our experiments. 

%% file: table/large_scale.tex
\begin{table*}[t!]
\begin{center}
\caption{
{
Objectives achieved from the MIS solvers on large-scale datasets, where the best objectives are marked in bold. Running times (in seconds) are provided in brackets, and the number of vertices is denoted by $N$. Out of budget (OB) indicates the runs violating the time budget (15 000 seconds) or the memory budgets (128GB RAM for CPU and 12GB VRAM for GPU).
}
}
\vspace{0.1in}
\label{tab:large_synth}
\scalebox{0.7}{
\begin{tabular}{
c 
c
c 
c 
c 
c @{\hspace{0.7cm}}
c 
c 
c @{\hspace{0.7cm}}
c 
c 
}
\toprule[1.2pt]
&
&
& \multicolumn{2}{c}{Classic}
&
& \multicolumn{2}{c}{SL-based}
& 
& \multicolumn{2}{c}{RL-based}
\\
\cmidrule[1.2pt]{4-5}
\cmidrule[1.2pt]{7-8} 
\cmidrule[1.2pt]{10-11}
\multicolumn{1}{c}{Type}
& \multicolumn{1}{c}{$N$} 
&
& \multicolumn{1}{c}{CPLEX}
& \multicolumn{1}{c}{KaMIS}
&
& \multicolumn{1}{c}{TGS}
& \multicolumn{1}{c}{TGS$^\dagger$}
&
& \multicolumn{1}{c}{LwD}
& \multicolumn{1}{c}{LwD$^\dagger$}
\\
\midrule[1.2pt]
\multirow{3.5}{*}{\shortstack[l]{BA}}
&500 000
&
&137 821 \small{(1129)\phantom{0}}
&228 123 \small{(1002)\phantom{0~}}
&
&227 701 \small{(344)\phantom{0}}
&228 733 \small{(6211)\phantom{0~}}
&
&228 803 \small{(45)\phantom{000}}
&\textbf{228 829} \small{(340)\phantom{00}}
\\
\addlinespace
&1 000 000 
&
&275 633 \small{(1098)\phantom{0}}
&457 541 \small{(2502)\phantom{0~}}
&
&455 354 \small{(620)\phantom{0}}
&457 073 \small{(10 484)\phantom{}}
&
&457 698 \small{(117)\phantom{00}}
&\textbf{457 752} \small{(651)\phantom{00}}
\\
\addlinespace
&2 000 000 
&
&551 767 \small{(1152)\phantom{0}}
&909 988 \small{(3637)\phantom{0~}}
&
&910 856 \small{(1016)\phantom{}}
&OB
&
&915 887 \small{(276)\phantom{00}}
&\textbf{915 968} \small{(1296)\phantom{0}}
\\
\midrule[1.2pt]
\multirow{3.5}{*}{\shortstack[l]{HK}} 
&500 000 
&
&\phantom{0}90 012 \small{(1428)\phantom{0}}
&175 153 \small{(1477)\phantom{0}}
&
&173 852 \small{(465)\phantom{0~}}
&176 253 \small{(777)\phantom{00~}}
&
&176 850 \small{(96)\phantom{000}}
&\textbf{177 143} \small{(519)\phantom{00}}
\\
\addlinespace
&1 000 000 
&
&179 326 \small{(1172)\phantom{0}}
&347 350 \small{(4463)\phantom{0~}}
&
&350 819 \small{(887)\phantom{0}}
&353 244 \small{(10 415)\phantom{}}
&
&353 504 \small{(248)\phantom{00}}
&\textbf{353 723} \small{(1151)\phantom{0}}
\\
\addlinespace
&2 000 000 
&
&OB 
&695 544 \small{(10 870)\phantom{}}
&
&OB
&OB
&
&706 975 \small{(648)}\phantom{00}
&\textbf{707 422} \small{(1601)\phantom{0}}
\\
\midrule[1.2pt]
\multirow{3.5}{*}{\shortstack[l]{WS}} 
&500 000 
&
&135 217 \small{(1424)\phantom{0}}
&\textbf{157 298} \small{(1002)\phantom{0}}
&
&153 190 \small{(354) \phantom{0}}
&155 230 \small{(10 172)\phantom{}}
&
&155 086 \small{(62)\phantom{000}}
&155 574 \small{(403)\phantom{00}}
\\
\addlinespace
&1 000 000 
&
&270 526 \small{(1093)\phantom{0}}
&303 810 \small{(1699)\phantom{0~}}
&
&308 065 \small{(1009)\phantom{}}
&OB
&
&310 308 \small{(144)\phantom{00}}
&\textbf{311 041} \small{(725)\phantom{00}}
\\
\addlinespace
&2 000 000 
&
&540 664 \small{(1159)\phantom{0}}
&603 502 \small{(4252)\phantom{0~}}
&
&611 057 \small{(1628)\phantom{}}
&OB
&
&620 615 \small{(345)\phantom{00}}
&\textbf{621 687} \small{(1388)\phantom{0}}
\\
\bottomrule[1.2pt]
\end{tabular}%
}
\end{center}
\end{table*}

%% file: figure/tradeoff.tex
\begin{figure*}[t]
\centering
\begin{subfigure}{.324\textwidth}
\centering
\includegraphics[width=1.0\textwidth]{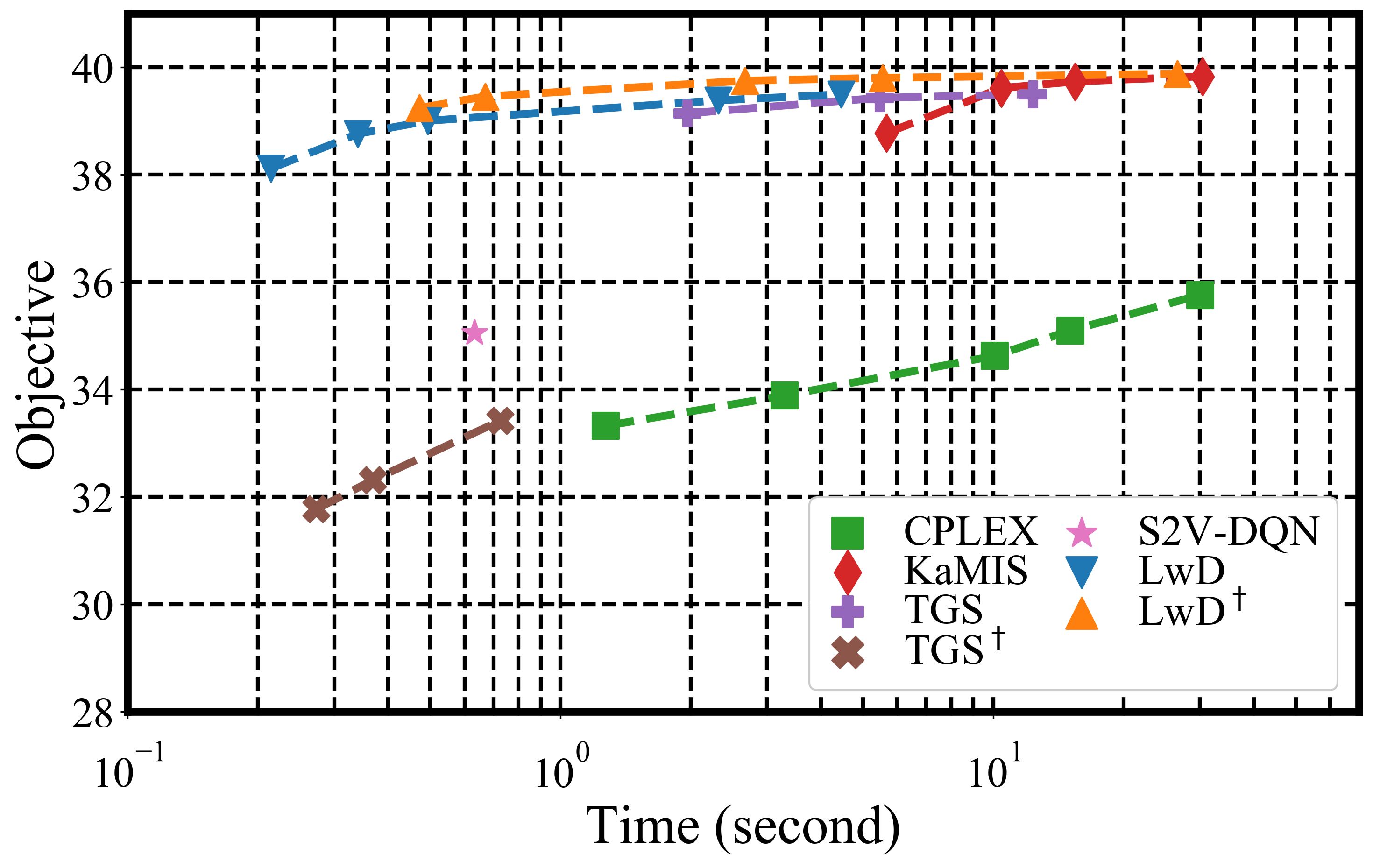}
\caption{ER-$(400, 500)$ dataset}
\end{subfigure}
\hspace{0.5in}
\begin{subfigure}{.3375\textwidth}
\centering
\includegraphics[width=1.0\textwidth]{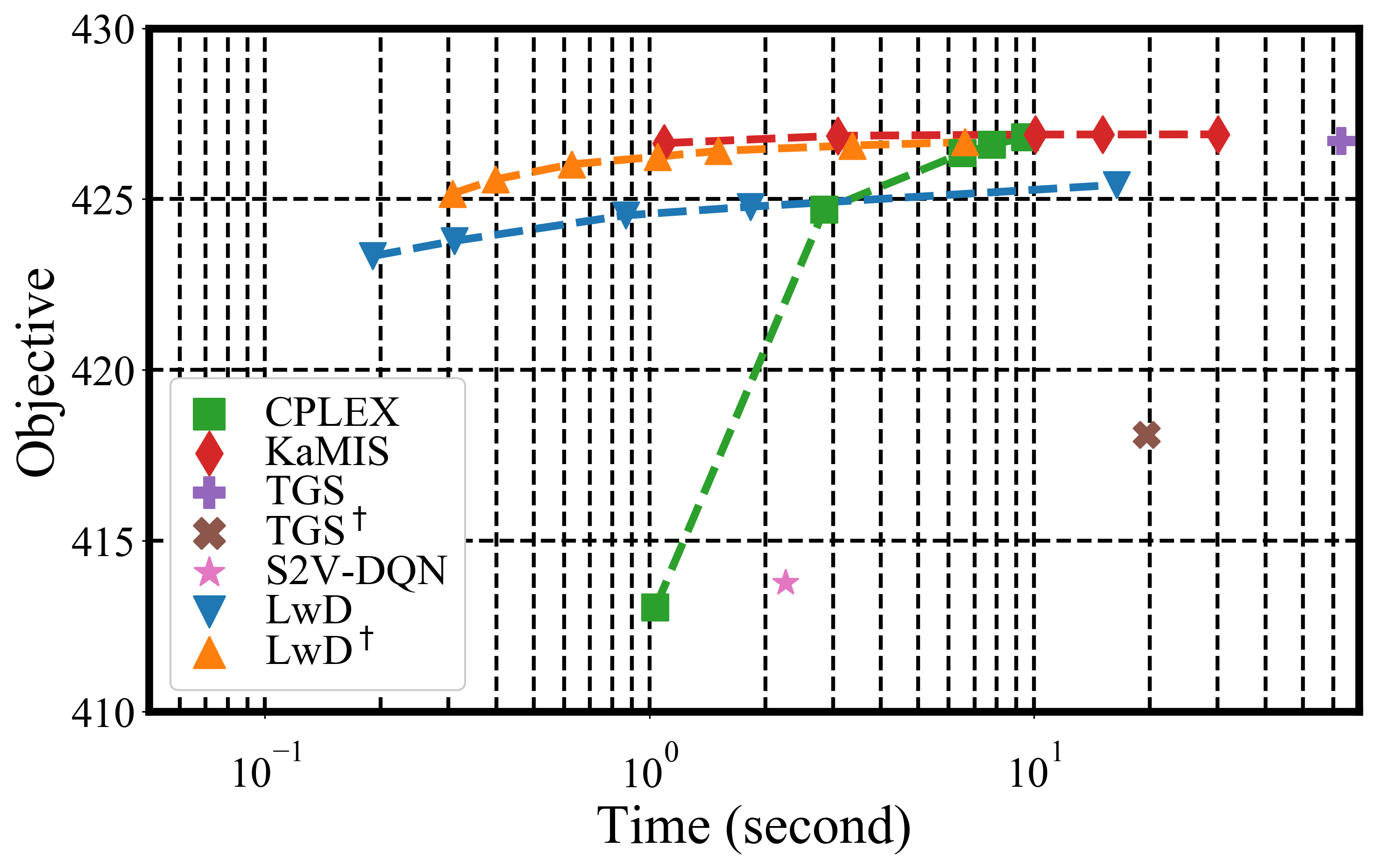}
\caption{SATLIB dataset}
\end{subfigure}
\caption{
Evaluation of trade-off between time and objective for the ER-$(400, 500)$ dataset (upper-left side is of better trade-off). 
}
\label{fig:tradeoff}
\end{figure*}

%% file: table/generalization.tex
\begin{table*}[t!]
\begin{center}
\caption{Approximation ratios of the deep-learning based MIS solvers on real-world graphs, i.e., Citation-\{Cora, Citeseer\}, Amazon-\{Photo, Computers\}, and Coauthor-\{CS, Physics\}, unseen during training. Best approximation ratios are marked in bold. Running times (in seconds) are provided in brackets, and the number of vertices is denoted by $N$. For computing the approximation ratios, We compute the optimal solutions from running the CPLEX solver without any time limit.}
\label{tab:gen_real}
\vspace{.1in}
\scalebox{0.7}{
\begin{tabular}{
c
c
c
c
c
r
r
r
}
\toprule[1.2pt]
&
&\multicolumn{2}{c}{SL-based}
&
&\multicolumn{3}{c}{RL-based}
\\
\cmidrule[1.2pt]{3-4}
\cmidrule[1.2pt]{6-8} 
\multicolumn{1}{c}{Type}
& \multicolumn{1}{c}{$N$} 
& \multicolumn{1}{c}{TGS}
& \multicolumn{1}{c}{TGS$^\dagger$}
& 
& \multicolumn{1}{c}{S2V-DQN}
& \multicolumn{1}{c}{LwD}
& \multicolumn{1}{c}{LwD$^\dagger$}
\\
\midrule[1.2pt]
Citation-Cora
&2708
&\textbf{1.00} \small{(4)\phantom{0}}
&\textbf{1.00} \small{(4)\phantom{000}}
&
&0.96 \small{(3)\phantom{000}}
&\textbf{1.00} \small{(3)\phantom{}}
&\textbf{1.00} \small{(3)\phantom{00}}
\\
\addlinespace
Citation-Citeseer 
&3327
&\textbf{1.00} \small{(3)\phantom{0}}
&\textbf{1.00} \small{(3)\phantom{000}}
&
&0.99 \small{(3)\phantom{000}}
&\textbf{1.00} \small{(2)\phantom{}}
&\textbf{1.00} \small{(4)\phantom{00}}
\\
\midrule[1.2pt]
Amazon-Photo
&7487
&0.99 \small{(9)\phantom{0}}
&\textbf{1.00} \small{(485)\phantom{0}}
&
&0.27 \small{(66)\phantom{00}}
&0.99 \small{(4)\phantom{}}
&\textbf{1.00} \small{(33)\phantom{0}}
\\
\addlinespace
Amazon-Computers
&13 381
&0.99 \small{(8)\phantom{0}}
&\textbf{1.00} \small{(823)\phantom{0}}
&
&0.26 \small{(236)\phantom{0}}
&0.99 \small{(3)\phantom{}}
&\textbf{1.00} \small{(101)\phantom{}}
\\ 
\midrule[1.2pt]
Coauthor-CS
&18 333
&0.99 \small{(17)\phantom{}}
&\textbf{1.00} \small{(80)\phantom{00}}
&
&0.88 \small{(197)\phantom{0}}
&\textbf{1.00} \small{(3)\phantom{}}
&\textbf{1.00} \small{(78)\phantom{0}}
\\
\addlinespace
Coauthor-Physics
&34 493
&0.98 \small{(52)\phantom{}}
&\textbf{1.00} \small{(1304)\phantom{}}
&
&0.19 \small{(1564)\phantom{}}
&0.98 \small{(9)\phantom{}}
&\textbf{1.00} \small{(186)\phantom{}}
\\
\bottomrule[1.2pt]
\end{tabular}
}
\end{center}
\end{table*}
%\footnotetext{We measure the approximation ratio using the optimal solutions from running the CPLEX solver without any time limit.}

%% file: figure/ablation.tex
\begin{figure*}[t]
    \centering
    \begin{subfigure}{0.3209\textwidth}
    \centering
    \includegraphics[width=0.9\textwidth]{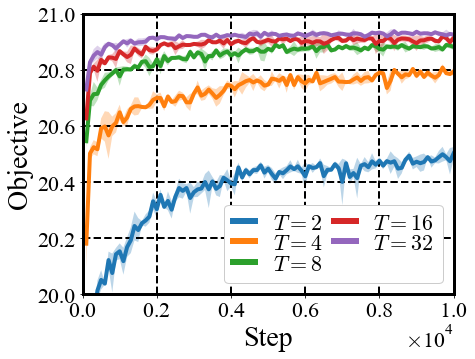}
    \caption{Performance with varying $T$}
    \label{fig:ablation_time}
    \end{subfigure}
    \hfill
    \begin{subfigure}{0.3209\textwidth}
    \centering
    \includegraphics[width=0.9\textwidth]{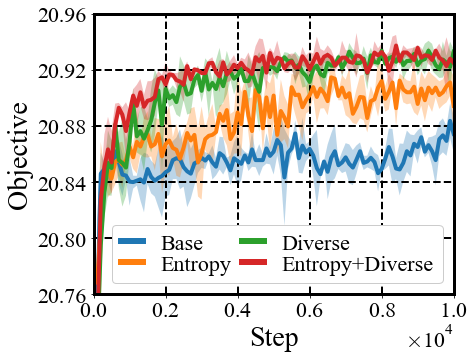}
    \caption{Contribution of each regularizers}
    \label{fig:ablation_hamming}
    \end{subfigure}
    \hfill
    \begin{subfigure}{0.3282\textwidth}
    \centering
    \includegraphics[width=0.9\textwidth]{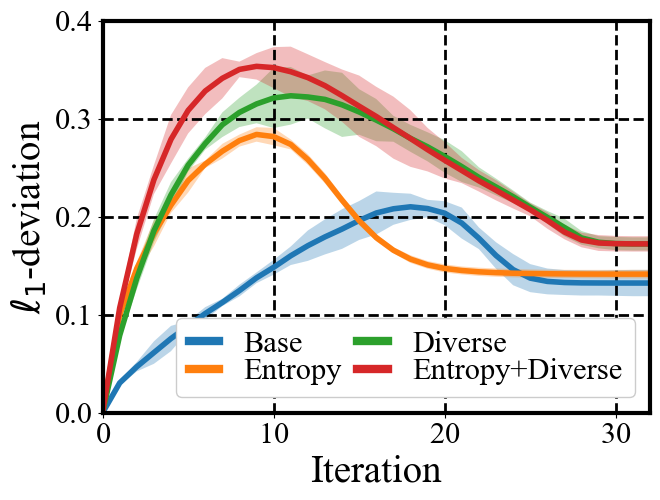}
    \caption{Deviation in intermediate stages}
    \label{fig:ablation_dev}
    \end{subfigure}
    \caption{
    Illustration of ablation studies done on ER-$(50, 100)$ dataset. 
    The solid line and shaded regions represent 
    the mean and standard deviation across 3 runs respectively
    Note that the standard deviation in (c) 
    was enlarged ten times for better visibility.}
\end{figure*}

%% file: table/other_comb.tex
\begin{table*}[t!]
\centering
\caption{
{
Objectives achieved by LwD and CPLEX for solving various combinatorial optimization problems 
on the ER datasets, where the best objectives are marked in bold. 
Running times (in seconds) for LwD are provided in brackets and CPLEX 
was run under time limit of $5$ seconds. 
The minimum and the maximum number of vertices are denoted by 
$N_{\mathtt{min}}$ and $N_{\mathtt{max}}$, respectively.
}
}
\vspace{0.1in}
\label{tab:mwis}
\scalebox{0.7}{
\begin{tabular}{
c 
c 
c
c 
c 
c@{\hspace{0.3cm}}
c 
c 
c@{\hspace{0.3cm}}
c 
c 
c@{\hspace{0.3cm}}
c 
c 
} 
\toprule[1.2pt]
& 
&
& \multicolumn{2}{c}{MWIS}
& 
& \multicolumn{2}{c}{PCMIS}
& 
& \multicolumn{2}{c}{MAXCUT}
& 
& \multicolumn{2}{c}{Ising}
\\
\cmidrule[1.2pt]{4-5}
\cmidrule[1.2pt]{7-8}
\cmidrule[1.2pt]{10-11}
\cmidrule[1.2pt]{13-14}
\multicolumn{1}{c}{Type}
&\multicolumn{1}{c}{$N_{\mathtt{min}}$}
&\multicolumn{1}{c}{$N_{\mathtt{max}}$}
&\multicolumn{1}{c}{CPLEX}
&\multicolumn{1}{c}{LwD}
&
&\multicolumn{1}{c}{CPLEX}
&\multicolumn{1}{c}{LwD}
&
&\multicolumn{1}{c}{CPLEX}
&\multicolumn{1}{c}{LwD}
&
&\multicolumn{1}{c}{CPLEX}
&\multicolumn{1}{c}{LwD}
\\
\midrule[1.2pt]
\multirow{3.5}{*}{ER}
&\phantom{0}50
&100
&\textbf{21.46}
&\textbf{21.46} \small{(0.16)}
&
&\textbf{22.64}
&21.77 \small{(0.01)}
&
&\textbf{251.6}
&240.4 \small{(0.01)}
&
&\phantom{0}108.8
&\textbf{151.9} \small{(0.01)}
\\
\addlinespace
&100
&200
&27.94
&\textbf{28.44} \small{(0.39)}
&
&27.12
&\textbf{29.56} \small{(0.11)}
&
&222.5
&\phantom{.}\textbf{1029} \small{(0.01)}
&
&\phantom{0}-479.7
&\textbf{406.8} \small{(0.02)}
\\
\addlinespace 
&400
&500
&34.29
&\textbf{39.21} \small{(0.54)}
&
&\phantom{0}4.56
&\textbf{38.39} \small{(0.42)}
&
&412.4
&\phantom{.}\textbf{7747} \small{(0.09)}
&
&\phantom{.}-14180
&\phantom{.}\textbf{1899} \small{(0.12)}
\\
\bottomrule[1.2pt]
\end{tabular}
}
\vspace{-.1in}
\end{table*}

%% file: appendix.tex
%\begin{center}{\bf {\LARGE Appendix:}}
%\end{center}
%\begin{center}{\bf {\Large Learning What to Defer for Maximum Independent Sets}}
%\end{center}

\section{Experimental Details}
\label{sec:app_detail}

\subsection{Implementation of LwD}
\label{sec:app_implementation}
In this section, we provide additional details for our implementation of the experiments. 

\textbf{GraphSAGE architecture.}
Each network consists of multiple layers $h_{n}$ with $n=1,\cdots, N$ where the $n$-th layer with weights $\mathbf{W}^{(n)}_{1}$ and $\mathbf{W}^{(n)}_{2}$ performs the following transformation on input $\mathbf{H}$:
\begin{align*}
h^{(n)}(\mathbf{H}) = \text{ReLU}\big(\mathbf{H}\mathbf{W}^{(n)}_{1}+\mathbf{D}^{-\frac12}\mathbf{A}\mathbf{D}^{-\frac12}\mathbf{H}\mathbf{W}^{(n)}_{2}\big). 
\end{align*}
Here $\mathbf{A}$ and $\mathbf{D}$ correspond to adjacency and degree matrix of the graph $\mathcal{G}$, respectively. At the final layer, the policy and value networks apply softmax function and graph readout function with sum pooling \citep{xu2018how} instead of $\text{ReLU}$ to generate actions and value estimates, respectively.

\textbf{Normalization of feature and reward.}
The iteration-index of MDP used for input of the policy and value networks was normalized by the maximum number of iterations. Both the cardinality reward (defined in Section \ref{subsec:mdp}) and the solution diversification reward (defined in Section \ref{subsec:diversification}) were normalized by maximum number of vertices in the corresponding dataset. 

\textbf{Hyperparameter.} 
Every hyperparameter was optimized on a per graph type basis and used across all sizes within each graph type. Throughout every experiment, the policy and the value networks were parameterized by graph convolutional network with $4$ layers and $128$ hidden dimensions. Every instance of the model was trained for 20 000 updates of proximal policy optimization \citepAP{schulman2017proximal}, based on the Adam optimizer with a learning rate of $0.0001$. The validation dataset was used for choosing the best performing model while using $10$ samples for evaluating the performance. Reward was not decayed throughout the episodes of the Markov decision process. Gradient norms were clipped by a value of $0.5$. We further provide details specific to each type of datasets in Table~\ref{tab:hyperparam}. For the compared baselines, we used the default hyperparameters provided in the respective codes. 

\input{table/hyperparam.tex}

\subsection{Implementation of Baselines}

\textbf{S2V-DQN.} 
We implement the S2V-DQN algorithm based on the code (written in C++) provided by the authors.%
\footnote{\url{https://github.com/Hanjun-Dai/graph_comb_opt}} For the synthetic graphs generated from ER, BA, HK, and SW models, S2V-DQN is unstable to be trained on graphs of size $(100, 200)$ and $(400, 500)$ without pre-training. Hence, we perform fine-tuning as mentioned in the original paper \citepAP{khalil2017learning}. For instance, when we train S2V-DQN on the ER-$(100, 200)$ datasets, we fine-tune the model trained on ER-$(50, 100)$. Next, for the ER-$(400, 500)$, we perform ``curriculum learning''; we first train S2V-DQN on the ER-$(50, 100)$ dataset, then fine-tune on the ER-$(100, 200)$, ER-$(200, 300)$, ER-$(300, 400)$ and ER-$(400, 500)$ in sequence. Finally, for training S2V-DQN on graphs with size larger than $500$, we were unable to train it on the raw graph under the available computational budget. Hence we train S2V-DQN on subgraphs sampled from the training graphs. To this end, we sample edges from the model uniformly at random without replacement, until the number of vertices reach $300$. Then we used the subgraph induced from the sampled vertices. 

\textbf{TGS.}
We use the official implementation and models provided by the authors.\footnote{\url{https://github.com/intel-isl/NPHard}} Unfortunately, the provided code runs out of memory for larger graphs since that they keep all of the intermediate solutions in the breadth-first search queue. For such cases, we modify the algorithm by discarding the oldest graph in the queue whenever the queue reaches its maximum size, i.e., ten times the number of required solutions for the problem.

\textbf{CPLEX.}
We use the CPLEX \citepAP{ilog2014cplex} provided in the official homepage.\footnote{\url{https://www.ibm.com/products/ilog-cplex-optimization-studio}} In order to optimize its performance under limited time, we set its emphasis parameter, i.e., \texttt{MIPEmphasisFeasibility}, to prefer higher objective over proof of optimality.

\textbf{KaMIS.}
We use KaMIS \citepAP{hespe2019wegotyoucovered} from its official hompage without modification.\footnote{\url{http://algo2.iti.kit.edu/kamis/}}

\subsection{Dataset Details}
\label{subsec:app_dataset}

In this section, we provide additional details on the datasets used for the experiments. 

\textbf{Synthetic datasets.} 
For the ER, BA, HK, and WS datasets, we train on graphs randomly generated on the fly and perform validation and evaluation on a fixed set of $1000$ graphs.

\textbf{SATLIB dataset.} 
The SATLIB dataset is a popular benchmark for evaluating SAT algorithms. We specifically use the synthetic problem instances from the category of random 3-SAT instances with controlled backbone size \citepAP{singer2000backbone}. Next, we describe the procedure for reducing the SAT instances to MIS instances. To this end, a vertex is added to the graph for each literal of the SAT instance. Then edges are added for each pair of vertices satisfying the following conditions: (a) that are in the same clause or (b) they correspond to the same literals with different signs. Consequently, the MIS in the resulting graph corresponds to the truth assignment to the optimal assignments of the SAT problem \citepAP{dasgupta2008algorithms}.

\textbf{PPI dataset.} 
The PPI dataset is the protein-protein-interaction dataset with vertices representing proteins and edges representing interactions between them. 

\textbf{REDDIT datasets.} 
The REDDIT-B, REDDIT-M-5K, and REDDIT-M-12K datasets are constructed from online discussion threads in reddit%
\footnote{\url{https://www.reddit.com/}} where vertices represent users and edges mean at least one of two users responded to the other user's comment.

\textbf{Autonomous system dataset.} 
The as-Caida dataset is a set of autonomous system graphs derived from a set of RouteViews BGP table snapshots \citepAP{leskovec2005graphs}. 

\textbf{Citation dataset.} 
The Cora and the Citeseer are networks constructed by vertices and edges representing documentation and citation links between them, respectively \citepAP{sen2008collective}.

\textbf{Amazon dataset.}
The Computers and Photo graphs are segmented from the Amazon co-purchase graph \citepAP{mcauley2015image}, where vertices correspond to goods and edges represent goods which are frequently purchased together.

\textbf{Coauthor dataset.} 
The CS and Physics graphs represent authors and the corresponding co-authorships by vertices and edges, respectively. It was collected from Microsoft Academic Graph from the KDD Cup 2016 challenge3.\footnote{\url{https://kddcup2016.azurewebsites.net/}}

We further provide the statistics of the datasets used in experiments corresponding to Table~\ref{tab:perf_synthetic} and~\ref{tab:gen_real} in Table~\ref{tab:dataset1} and~\ref{tab:dataset2}, respectively.
\input{table/dataset.tex}

%\newpage
%\newpage
\clearpage
\section{Details of Other Combinatorial Optimizations}
\label{sec:app_variant}

\subsection{Maximum Weighted Indpendent Set Problem} 
First, we describe the maximum weighted independent set (MWIS) problem \citepAP{balas1986finding}. Consider a graph $\mathcal{G} = (\mathcal{V}, \mathcal{E})$ associated with positive weight function $w:\mathcal{V} \rightarrow \mathbb{R}^{+}$. The goal of the MWIS problem is to find the independent set $\mathcal{I} \subseteq \mathcal{V}$ where the total sum of weight $\sum_{i \in \mathcal{I}}w(i)$ is maximum. In order to apply the LwD framework to the MWIS problem, we include the weight of each vertex as its feature to the policy network and modify the reward function by the increase in weight of included vertices, i.e., $R(\bm{s}, \bm{s}^{\prime}) = \sum_{i\in\mathcal{V}_{\ast} \setminus \mathcal{V}^{\prime}_{\ast}} s_{i}^{\prime}w(i)$. We sample the weights of each vertices from a normal distribution with mean and standard deviation fixed to $1.0$ and $0.1$, respectively. 

\subsection{Prize Collecting Maximum Independent Set Problem}
The prize collecting maximum independent set (PCMIS) problem is an instance of the generalized minimum vertex cover problem \citepAP{hassin2006minimum}. To describe this problem, one may consider a graph $\mathcal{G} = (\mathcal{V}, \mathcal{E})$ and a subset of vertices $\mathcal{I}\subseteq \mathcal{V}$. The PCMIS problem is associated with the following the ``prize'' function $\mathnormal{f}$ to maximize: 
\begin{equation*}
\mathnormal{f}(\mathcal{I}) := |\mathcal{I}| - \lambda |\{\{i, j\}: i, j\in\mathcal{I}, i \neq j\}|,
\end{equation*} 
where $\lambda > 0$ is the penalty function for including two adjacent vertices. We set $\lambda=0.5$ in the experiments. Such a problem could be interpreted as relaxing the hard constraints on independent set to a penalty function in the MIS problem. Especially, one can examine that optimal solution of the PCMIS problem becomes the maximum independent set when $\lambda > 1$. For applying the LwD framework on the PCMIS problem, we remove the clean-up phase in the transition function of MDP and modify the reward function $R(\bm{s}, \bm{s}^{\prime})$ as the increase in prize function at each iteration.

\iffalse
, expressed as follows: 
\begin{equation*}
    R(\bm{s}, \bm{s}^{\prime}) := 
    \sum_{i\in\mathcal{V}_{\ast} \setminus \mathcal{V}^{\prime}_{\ast}} 
    \bigg( 
    s_{i}^{\prime}
    - \lambda 
    \sum_{j \in \mathcal{V}_{\ast} \setminus \mathcal{V}^{\prime}_{\ast} \setminus \{i\}}
    \frac12 s_{i}^{\prime}s_{j}^{\prime}
    + 
    \lambda 
    \sum_{j \in \mathcal{V} \setminus \mathcal{V}_{\ast}}
    s_{i}^{\prime}s_{j}^{\prime}
    \bigg).
\end{equation*}
\fi

\subsection{Maximum Cut Problem}
Next, we introduce the maximum cut (MAXCUT) problem \citep{garey1979computers}. Given a graph $\mathcal{G}=(\mathcal{V}, \mathcal{E})$, goal of the MAXCUT problem is on finding a subset of vertices $\mathcal{I}$ that maximize the following objective:
\begin{equation*}
    \mathnormal{f}(\mathcal{I}) := |\{i, j\}: i \in \mathcal{I}, j \in \mathcal{V}\setminus\mathcal{I}|,
\end{equation*}
which corresponds to the number of edges that form a ``cut'' between $\mathcal{I}$ and $\mathcal{V}\setminus \mathcal{I}$. To apply LwD, we remove the clean-up phase in the transition function of our MDP and set the reward function $R(\bm{s}, \bm{s}^{\prime})$ as the increase in the objective function at each iteration. 

\subsection{Maximum-a-posteriori Inference Problem on the Ising Model}
Finally, we describe the maximum-a-posteriori (MAP) inference problem on the anti-ferromagnetic Ising model \citepAP{onsager1944crystal}. Given a graph $\mathcal{G} = (\mathcal{V}, \mathcal{E})$, the probability distribution of the Ising model is described as $p(\bm{s}) := \frac1Z \exp{\big(\phi(\bm{s})\big)}$, where $\bm{s}$ is the random variable, $Z$ is the normalization constant, and the function $\phi$ is the objective to maximize. To be specific, $\phi$ is defined as follows:
\begin{align*}
\phi(\bm{s}) :&= \gamma \sum_{i \in \mathcal{V}} \mathnormal{f}_{1}(i)+ \beta \sum_{\{i, j\}\in\mathcal{E}} \mathnormal{f}_{2}(i,j), \\
\mathnormal{f}_{1}(i) &= \begin{cases}-1 \quad &\text{if}~s_{i}=0 \\
1 \quad &\text{if}~s_{i}=1\end{cases},\qquad \mathnormal{f}_{2}(i, j) = \begin{cases}-1 \quad &\text{if}~s_{i}=s_{j} \\
1 \quad &\text{if}~s_{i} \neq s_{j}
\end{cases},
\end{align*}
Here, $\beta$ and $\gamma$ are called interaction and magnetic field parameters, respectively. Furthermore, $\bm{1}_{\mathcal{A}}$ is an indicator function for set of vertices $\mathcal{A}$ and $\mathcal{A}_{k} = \{i: i\in \mathcal{V}, s_{i} = k\}$ for $k= 0, 1$. In order to solve the MAP inference problem on the Ising model, we remove the clean-up phase in the transition function as in the PCMIS problem and modify the reward function $R(\bm{s}, \bm{s}^{\prime})$ as the increase in objective function at each iteration.
In our experiments, we set $\beta=1.0, \gamma=1.0$ and report $\phi(\bm{s})$ as the objective to maximize.

\iffalse
expressed as follows:
\begin{equation*}
R(\bm{s}, \bm{s}^{\prime}):= \sum_{i \in \mathcal{V}_{\ast} \setminus \mathcal{V}_{\ast}^{\prime}} \bigg(\gamma \mathnormal{f}_{1}(i)+ \beta \sum_{j \in \mathcal{V}_{\ast} \setminus \mathcal{V}^{\prime}_{\ast} \setminus \{i\}}\frac12\mathnormal{f}_{2}(i, j) + \beta\sum_{j \in \mathcal{V} \setminus \mathcal{V}_{\ast}}\mathnormal{f}_{2}(i,j)\bigg).
\end{equation*}
\fi

\newpage
\section{Graphical Illustration of LwD}
\label{sec:app_graph}
\input{figure/audp.tex}

%% file: table/hyperparam.tex
\begin{table}[h]
\begin{center}
\caption{
Choice of hyperparameters for the experiments on 
performance evaluation. The REDDIT column indicates hyperparameters used for the REDDIT-B, REDDIT-M-5K, and REDDIT-M-12K datasets. 
}
\label{tab:hyperparam}
\resizebox{\columnwidth}{!}{
\begin{tabular}{lcccccc}
\toprule
Parameters
&
ER, BA, HK, WS
& 
SATLIB 
& 
PPI 
& 
REDDIT
& 
as-Caida \\
\midrule
Maximum iterations per episode
& 32 
& 128 
& 128
& 64
& 128
\\
\midrule
Number of unrolling iteration
& 32 
& 128 
& 128
& 64
& 128
\\
\midrule
Number of environments per batch (graph instances)
& 32
& 32
& 10
& 64
& 1
\\\midrule
Batch size for gradient step
& 16
& 8
& 8
& 16 
& 8
\\
\midrule
Number of gradient steps per update
& 4
& 8
& 8
& 16
& 8 
\\
\midrule
Solution diversity reward coefficient 
& 0.1
& 0.01
& 0.1
& 0.1
& 0.1 
\\
\midrule
Maximum entropy coefficient
& 0.1
& 0.01 
& 0.001
& 0.0
& 0.1 
\\
\bottomrule
\end{tabular}
}
\end{center}
\end{table}

%% file: table/dataset.tex
\begin{table}[H]
\begin{center}
\caption{
Number of nodes, edges and graphs for SATLIB, PPI, REDDIT, and as-Caida datasets used in Table \ref{tab:perf_synthetic}. 
Number of graphs is expressed as a tuple of 
the numbers of training, validation and test graphs, respectively.
}
\vspace{.1in}
\label{tab:dataset1}
\begin{tabular}{lccc}
\toprule
Dataset
&
Number of nodes
&
Number of edges
& 
Number of graphs
\\
\midrule
SATLIB
& (1209, 1347)
& (4696, 6065)
& (38 000, 1000, 1000)
\\
\midrule
PPI
& (591, 3480)
& (3854, 53 377)
& (20, 2, 2)
\\
\midrule
REDDIT (BINARY)
& (6, 3782)
& (4, 4071)
& (1600, 200, 200)
\\
\midrule
REDDIT (MULTI-5K)
& (22, 3648)
& (21, 4783)
& (4001, 499, 499)
\\
\midrule
REDDIT (MULTI-12K)
& (2, 3782)
& (1, 5171)
& (9545, 1192, 1192)
\\
\midrule
as-Caida
& (8020, 26 475)
& (36 406, 106 762)
& (108, 12, 12)
\\
\bottomrule
\end{tabular}
%}
\end{center}
\end{table}

\begin{table}[H]
\caption{
Number of nodes and edges for each dataset used in the Table \ref{tab:gen_real}. 
}
\label{tab:dataset2}
\vspace{.1in}
\begin{center}
\begin{tabular}{lcc}
\toprule
Dataset
&
Number of nodes
&
Number of edges
\\
\midrule
Citeseer
& 3327
& 3668
\\
\midrule
Cora
& 2708
& 5069
\\
\midrule
Coauthor CS
& 18 333
& 81 894
\\
\midrule
Coauthor Physics
& 34 493
& 247 962
\\
\midrule
Amazon Computers
& 13 381
& 245 778
\\
\midrule
Amazon Photo 
& 7487
& 119 043
\\
\bottomrule
\end{tabular}
\end{center}
\end{table}

%% file: figure/audp.tex
\begin{figure}[hbt]
    \centering
    \includegraphics[width=0.60\textwidth]{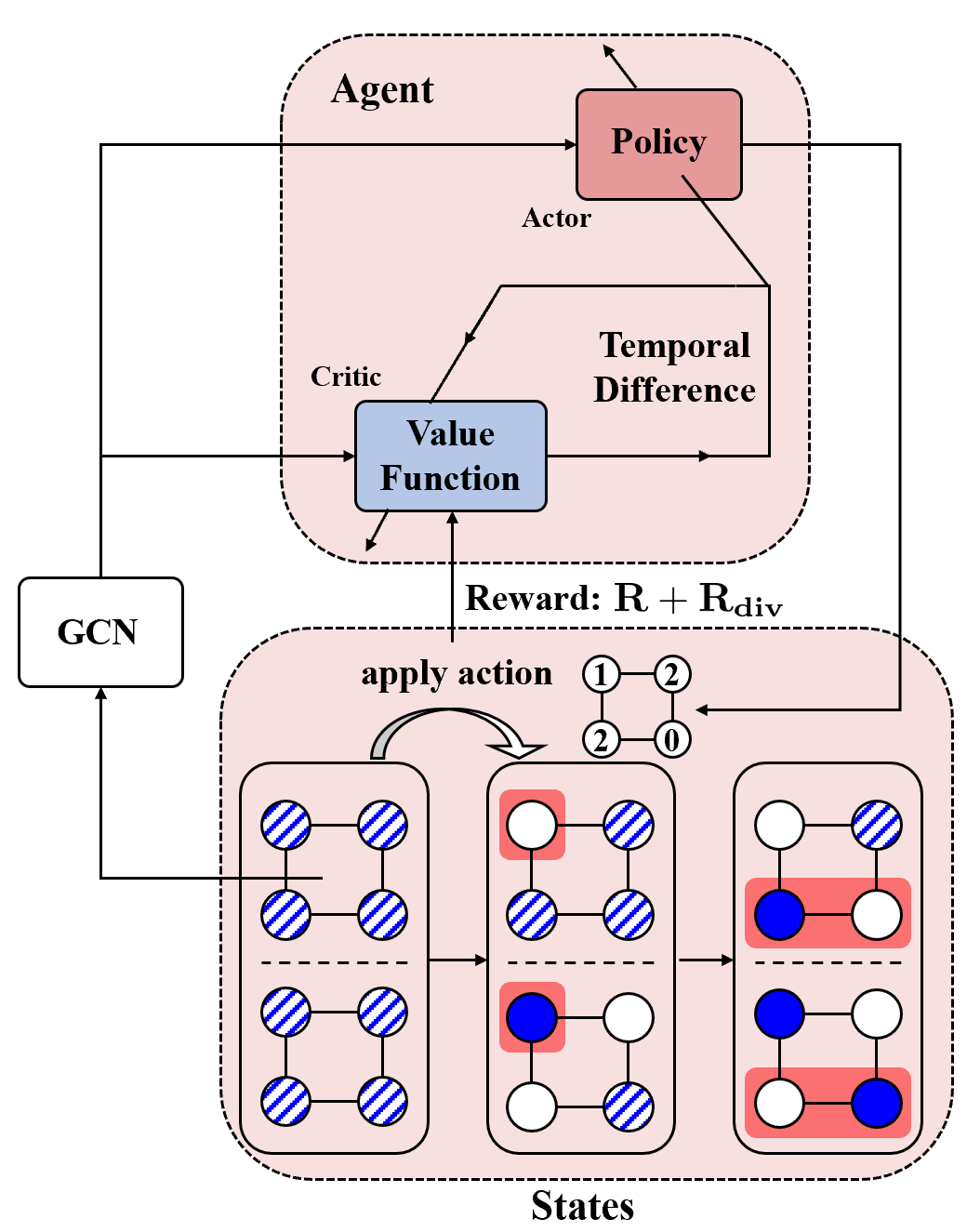}
    \caption{Illustration of the overall learning what to defer (LwD) framework.}
    \label{fig:audp}
\end{figure}